\documentclass[letterpaper, 10 pt, conference]{Settings/ieeeconf}  % Comment this line out if you need a4paper

\IEEEoverridecommandlockouts                              % This command is only needed if 
\usepackage{footnote}
\usepackage{amsmath} % assumes amsmath package installed
\usepackage{amssymb}  % assumes amsmath package installed
\usepackage{booktabs}
\usepackage{color}
\usepackage{xcolor}
\usepackage{multirow}

\usepackage[pdftex]{graphicx}
\usepackage{siunitx}
\usepackage{cite}

\usepackage{hyperref}
\usepackage{subcaption}
\usepackage{glossaries}

\newacronym{HRI}{HRI}{human-robot interaction}
\newacronym{EBC}{EBC}{Expectation-Based Control}
\newacronym{MPC}{MPC}{Model Predictive Control}
\newacronym[longplural={degrees of freedom},shortplural={DoFs}]{DoF}{DoF}{degree of freedom}
\newacronym{SMU}{SMU}{Safe Motion Unit}
\newacronym{SSM}{SSM}{speed and separation monitoring}
\newacronym{PFL}{PFL}{power and force limiting}
\newacronym[longplural={inertial measurement units},shortplural={IMUs}]{IMU}{IMU}{inertial measurement unit}
\newacronym[longplural={evasive motions},shortplural={EMs}]{EM}{EM}{evasive motion}
\newacronym[longplural={Expectable Motion Units},shortplural={EMUs}]{EMU}{EMU}{Expectable Motion Unit}
\newacronym{FE}{FE}{Franka Emika}
\newacronym{IMO}{IMO}{human involuntary motion occurrence}
\newacronym{TUI}{TUI}{Technology Usage Inventory}
\newacronym{PS}{PS}{perceived safety}
\newacronym{GS}{GS}{Godspeed}

\definecolor{excelblue}{HTML}{4472C4} 
\definecolor{NewBlue}{HTML}{4472C4} 
\definecolor{NewOrange}{HTML}{ED7D31} 
\definecolor{NewGreen}{HTML}{00B050} 

\usepackage{stackengine}
\newcolumntype{P}[1]{>{\centering\arraybackslash}p{#1}}

\usepackage{fancyhdr}
\input{Settings/abkuerzung.sty}
\graphicspath{{graphics/}}
\DeclareGraphicsExtensions{.pdf,.jpeg,.png}

\renewcommand{\baselinestretch}{1}

\def\mytitle{Energy-Based Injury Protection Database: Including Shearing Contact Thresholds for Hand and Finger Using Porcine Surrogates}
\title{\LARGE \bf \mytitle}
\def\myauthor{ Robin Jeanne Kirschner$^{1,2}$, Anna Huber$^{1}$, Carina M. Micheler$^{2,3}$,  \\Dirk Müller$^{2}$, Nader Rajaei$^{1}$, Rainer Burgkart$^{2}$ and Sami Haddadin$^{4}$}
\def\mythanks{$^1$ Munich Institute of Robotics and Machine Intelligence, Technical University of Munich, 80992 Munich, Germany\newline 
$^2$ TUM Hospital, Department of Orthopaedics and Sports Orthopaedics, 81675 Munich, Germany\newline
$^3$ Institute for Machine Tools and Industrial Management, TUM School of Engineering and Design, Technical University of Munich, 85748 Garching near Munich, Germany\newline
$^4$  Mohamed Bin Zayed University of Artificial Intelligence, Masdar City, Abu Dhabi, United Arab Emirates\newline
Corresponding author:
{\href{mailto:robin-jeanne.kirschner@tum.de}{\tt\small robin-jeanne.kirschner@tum.de}}
}

\author{\myauthor% <-this % stops a space
\thanks{}% <-this % stops a space
\thanks{\mythanks}%
}

\hypersetup{ % sets Metadata of the pdf file
pdftitle={\mytitle},
pdfauthor={\myauthor},
}

\DeclareUnicodeCharacter{2009}{\,}

\begin{document}

\maketitle
\thispagestyle{empty}
\pagestyle{empty}
\thispagestyle{empty}
\thispagestyle{fancy} %FOR ARXIV
\pagestyle{empty}
\pagestyle{fancy} %FOR ARXIV

%%%%%%%%%%%%%%%%%%%%%%%%%%%%%%%%%%%%%%%%%%%%%%%%%%%%%%%%%%%%%%%%%%%%%%%%%%%%%%%%
\begin{abstract}

While robotics research continues to propose strategies for collision avoidance in human-robot interaction, the reality of constrained environments and future humanoid systems makes contact inevitable. To mitigate injury risks, energy-constraining control approaches are commonly used, often relying on safety thresholds derived from blunt impact data in EN ISO 10218-2:2025. However, this dataset does not extend to edged or pointed collisions. Without scalable, clinically grounded datasets covering diverse contact scenarios, safety validation remains limited. Previous studies have laid the groundwork by assessing surrogate-based velocity and mass limits across various geometries, focusing on perpendicular impacts. This study expands those datasets by including shearing contact scenarios in unconstrained collisions, revealing that collision angle significantly affects injury outcomes. Notably, unconstrained shearing contacts result in fewer injuries than perpendicular ones. By reevaluating all prior porcine surrogate data, we establish energy thresholds across geometries and contact types, forming the first energy-based Injury Protection Database. This enables the development of meaningful energy-limiting controllers that ensure safety across a wide range of realistic collision events.

\end{abstract}
%important because robot will be part of normal environments with non-expert users
%%%%%%%%%%%%%%%%%%%%%%%%%%%%%%%%%%%%%%%%%%%%%%%%%%%%%%%%%%%%%%%%%%%%%%%%%%%%%%%%
%\printinunitsof{cm}\prntlen{\textwidth}
%\printinunitsof{cm}\prntlen{\columnwidth}

%%%%%%%%%%%%%%%%%%%%%%%%%%%%%%%%%%%%%%%%%%%%%%%%%%%%%%%%%%%%%%%%%%%%%%%%%%%%%%%%
%%%%%%%%%%%%%%%%%%%Introduction%%%%%%%%%%%%%%%%%%%%%%%%%%%%%%%%%%%%%%%%%%%%%%%%%
%%%%%%%%%%%%%%%%%%%%%%%%%%%%%%%%%%%%%%%%%%%%%%%%%%%%%%%%%%%%%%%%%%%%%%%%%%%%%%%%
\glsresetall %Macht dass Abkürzungen die hier vor schon benutzt wurden hier nach erneut eingeführt werden.
\section{Introduction}

Energy-limiting strategies are among the most promising approaches for enabling safe human-robot interaction (HRI) in dynamic, autonomous environments \cite{lachner2021energy, Thumm2025, Haddadin.2012}. However, despite extensive research in injury biomechanics and accident analysis, energy-based injury thresholds relevant to HRI remain scarce \cite{Bessler_2021, lachner2021energy}. Most available data stem from clinical experience or severe injury cases in automotive crash scenarios \cite{Fujikawa_2013}, which primarily concern head, chest, or neck trauma—less relevant for HRI, where hand and finger injuries are more likely and should be minimized \cite{DeutscheGesetzlicheUnfallversicherung2024, Sanders2024}.

 \begin{figure}[th]
	
		\includegraphics[width=1\linewidth]{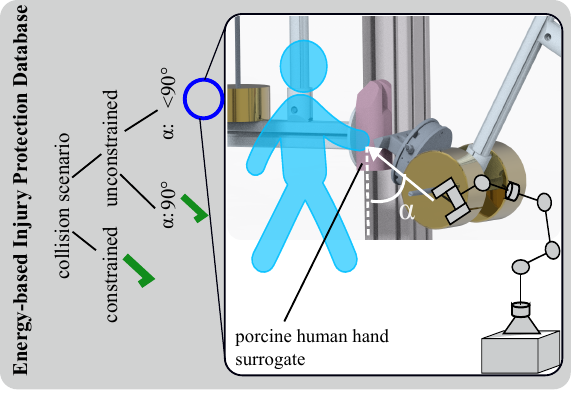}
	\caption{ Shearing collisions, common in real-world contact scenarios, are not yet covered by existing energy-based safety thresholds. To investigate injury severity, the robot is modeled as a moving mass with defined impact geometry and collision angle. Surrogate experiments provide initial energy limits based on contact geometry, enabling estimation of safe energy densities for non-perpendicular impacts and adding it to the Injury Protection Database. }
	\label{fig:specimen}
	
	\vspace{-5mm}
	\label{fig:intro}
\end{figure}

To address this gap, surrogate studies have investigated low-severity injuries such as skin contusions \cite{Sugiura_2022, Lesueur_2025, Haddadin.2012}, pain onset from blunt impacts \cite{Yamada_1996, Han_2021, Povse_2010}, bone fractures \cite{Hohendorff20112158, Kent_2008}, and abrasions caused by wearable robot devices \cite{mao2017characteristics}. Yet, structured datasets covering a range of injury types and severity, which would be relevant for appropriate risk assessment of potential contact scenarios, especially for contacts with sharp or pointed geometries, are still missing.

Moreover, existing safety testing frameworks rely on conservative assumptions, such as zero pain tolerance and fixed contact areas (e.g., 1~cm$^2$) \cite{iso_10218-2}, limiting flexibility in risk assessment. Force- and pressure-based thresholds require precise contact modeling and experimental assessment of the occurring forces, which is impractical and unnecessarily complex to apply in real-world settings.

%\cor{As illustrated in Fig.~\ref{fig:intro}, collision scenarios can be modeled by robot mass and velocity model unconstrained collision scenarios using defined impact geometries, effective masses, and robot velocities to systematically investigate injury severity and occurrence. Without scalable, clinically grounded datasets, safety validation for robots in human-centric environments remains fundamentally constrained.}

Previous research focused on assessing injury limits based on mass and velocity information using pig feet as a surrogate for human hands \cite{Kirschner_2024_Towards, Kirschner_2024_unconst}. These are the fundamental bases for a comprehensive energy-based Injury Protection Database, allowing human injury estimation based on surrogate studies. 

In this paper, we obtained original data from previous studies and reassessed these to provide an overview of energy limits \cite{Kirschner_2024_Towards, Kirschner_2024_unconst}. Additionally, we assess for unconstrained contacts whether the severity of harm may change in the likely case of non-perpendicular contact, but shearing along the body part instead. For this assessment, we present the measurement procedure and experimental setup to structurally investigate injury severity caused in unconstrained shearing contacts. Using this setup, we conduct a surrogate study with pig front paws as a substitute for human hands with sharp wedged, edged, and sheet-shaped impactors, two different effective masses of human body parts for collision with human arms, five effective robot masses, and three robot velocities. Lastly, we compare the obtained injury data to previous investigations on injuries occurring in normal unconstrained impact scenarios and provide an overview of the energy boundaries to be considered in different collision scenarios in human-robot interaction based on these surrogate studies.

This paper is structured as follows. Sec.\ref{sec:State of the Art} summarizes the foundational experiments to this research and the resulting problem statement. Sec.\ref{sec:methodology} motivates and describes the conducted injury experiments. The results are presented in Sec.\ref{sec:results}. Sec.\ref{sec:Discussion} then discusses and outlines exploitation and limitations and we conclude the paper in Sec.\ref{sec:Conclusion}.

%%%%%%%%%%%%%%%%%%%%%%%%%%%%%%%%%%%%%%%%%%%%%%%%%%%%%%%%%%%%%%%%%%%%%%%%%%%%%%%%
%%%%%%%%%%%%%%%%%%%%%%%%%%State of the Art%%%%%%%%%%%%%%%%%%%%%%%%%%%%%%%%%%%%%%
%%%%%%%%%%%%%%%%%%%%%%%%%%%%%%%%%%%%%%%%%%%%%%%%%%%%%%%%%%%%%%%%%%%%%%%%%%%%%%%%

\section{Previous Work and Problem Statement}\label{sec:State of the Art}
Here, we summarize and compare the two experimental approaches used to investigate collision scenarios on surrogate biological specimens under constrained and unconstrained conditions, which we are extending and that serve as a basis for this study, being ~\cite{Kirschner_2024_Towards} and ~\cite{Kirschner_2024_unconst}. 

Both studies focused on deriving information about injuries that occur at initial \textit{perpendicular} collision with certain masses and velocities using varying impact geometries. Both employed the three impact geometries: 

\begin{itemize}
    \item wedge with \SI{90}{^\circ} angle,
    \item edge with \SI{90}{^\circ} angle,
    \item \SI{1.6}{mm} thick sheet.
\end{itemize}  

Pig dew claws from the front paw were consistently used to represent human fingers. %\cite{lunney2021importance}.
Additionally, in the second study on unconstrained collisions, the back of the pig´s front paw simulated the human back of the hand.

\subsubsection{Constrained Collision Scenarios}

In the first study, a drop test device was utilized to simulate impacts with a constrained human hand or finger under varying conditions, including masses from \SI{0.5}{} to \SI{8}{kg} and velocities between \SI{0.20}{} and \SI{2.0}{m/s}. The mass-velocity limits in this study were derived from the contact speed based on the elevation height of the falling mass (verified by a calibration curve) and the dropped mass. A Kistler Type 9331C \cite{kistler9331c} piezo-electric force sensor with a measurement range of up to \SI{24}{kN} was utilized to provide also initial results on forces causing different types of injuries.

\subsubsection{Unconstrained, Perpendicular Collision Scenarios}

In ~\cite{Kirschner_2024_unconst}, a pendulum-based test setup was applied. The pendulum was elevated using a motor, belt, and winch mechanism to model the mass and velocity of a robotic collision. The aforementioned impactors were mounted on top of the pendulum to simulate the geometry in contact conditions. Force measurements in this study were obtained using a K6D40 force sensor with a range of up to \SI{500}{N}~\cite{force_datasheet}. The mass-velocity limits provided here are given based on the measured contact speed and effective pendulum mass.

Both studies focus on the consequence of perpendicular contact with these geometries, but they do not cover the question, whether shearing contact, at different impact angles should be considered separately. They also supply masses and velocities causing certain injury types separately but not yet the energy thresholds resulting from them. 

\subsubsection{Injury dynamics}

% With a focus on fingers and hands, injury is caused by the energy density acting on the human (or porcine) skin. This causes stress and strain on the different tissue levels, causing the skin or underlying tissue to react \cite{whittle2008biomechanical, elsafty2024insights,shergold2006uniaxial}. If the human body part can recoil in the same direction of motion as the hit, the initial impact is mostly shaped by the energy in the initial contact surface. If it cannot recoil, in addition to the initial energy density, the internal dynamics of the finger induce shearing forces as the skin on the stiffer bone tissue (fingers) tries to escape and moves on top of the bone. And lastly, if it can recoil, but does not recoil in the same direction as the contact, an additional shearing force is caused on top of the skin. 
% This is specifically relevant in unconstrained collision situations, as undesired contact in unconstrained situations is generally more likely than in unconstrained conditions and the motion direction and recoiling direction of the body part is rarely ever parallel, unless the collision is expected by the human and caused intentionally.

% Consequently, to provide an overview of relevant energy thresholds for shaping safety-informed energy-control, we need to ask whether shearing contact situations in unconstrained contacts may carry greater injury potential and then map all three studies' results into the energy domain.
\begin{figure}[t]
\centering
\includegraphics[width=1\linewidth]{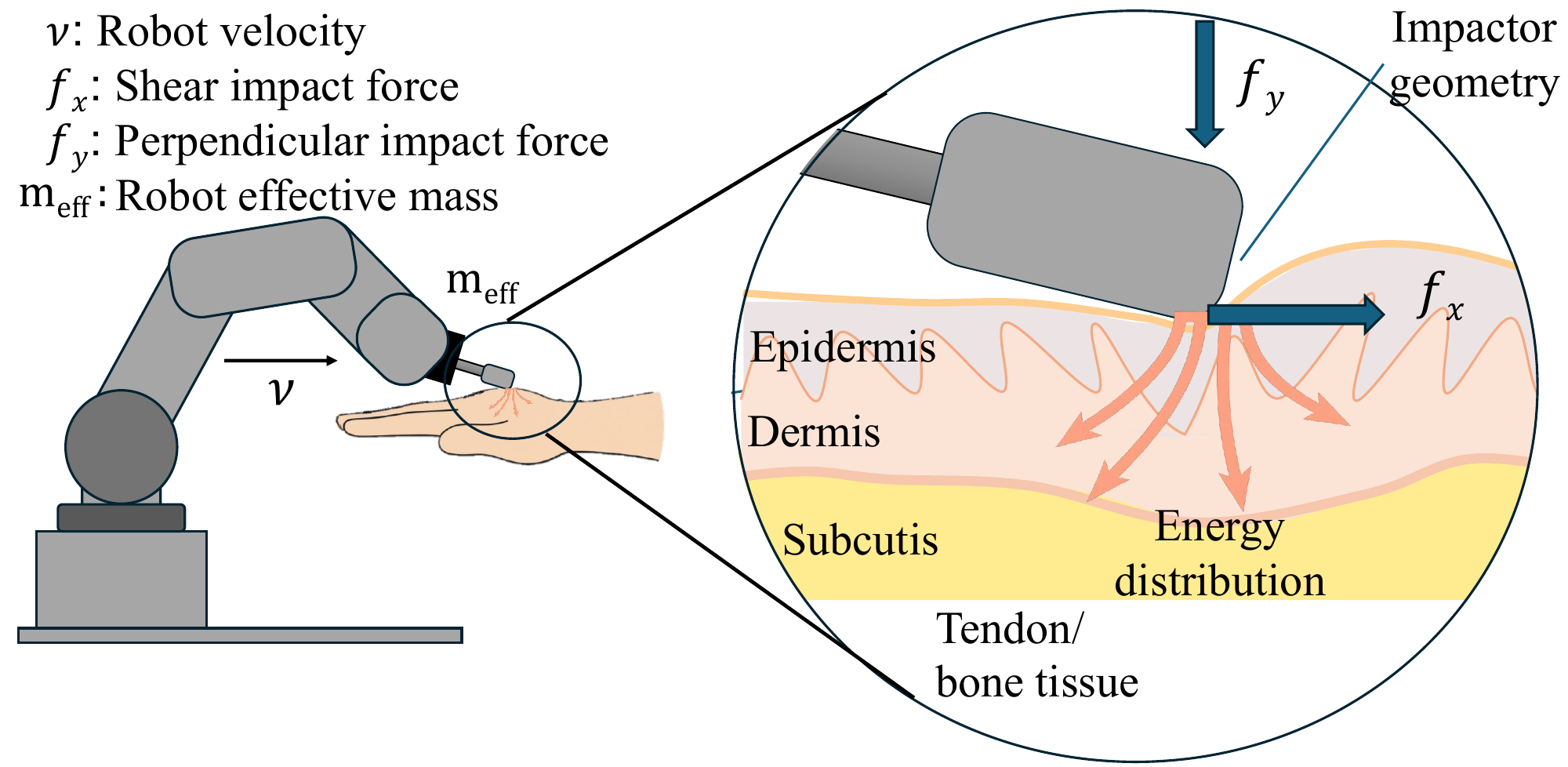}
\caption{Schematic of the shearing contact and energy distribution as well as skin deformation. The individual layers of tissue can be seen here qualitatively specifically on the back of the hand. The muscle tissue can be disregarded.}
\vspace{-0.2cm}
\label{fig:skin}
\vspace{-4mm}
\end{figure}
% Hinweis zu der Abbildung: Vom Verhältnis ist die Dermis die dickste Schicht und die Epidermis und Subcutis  dagegen sehr dünn. Die Subcutis fehlt hier aktuell noch und kann auch am Handrücken auftreten. Ich würde hier die gelbe Schicht als Subcutis bezeichnen und unten den weißen Teil etwas vergrößern und dann als Tendon/bone tissue bennen. 
% Zusätzlich ist im Allgemeinen zw. Subcutis und Knochen noch das Muskelgewebe. Wenn du hier explizit die Schichten im Handrücken zeigen möchtest, dann kannst du es so lassen. Ich würde dann nur in der Bildunterschrift noch ergänzen: The individual layers of tissue can be seen here specifically on the back of the hand. The muscle tissue can be disregarded. 

With a focus on fingers and hands, injury is caused by the energy density acting on human (or porcine) skin. This induces stress and strain across different tissue layers, causing the skin or underlying tissue to deform \cite{whittle2008biomechanical} as indicated in Fig. \ref{fig:skin}. The relation between the shearing force $f_\mathrm{x}$ and normal force,$f_\mathrm{y}$, is crucial for the injury pattern \cite{mao2017characteristics}. If the body part can recoil in the same direction as the impact, $f_\mathrm{x}$ in Fig. \ref{fig:skin} becomes negligibly small and the injury is primarily determined by the energy at the initial contact surface. If it cannot recoil $f_\mathrm{x}$ increases and internal dynamics induce additional shearing forces, as the skin and soft tissue moves relative to the underlying stiffer tendon and bone tissue and uneven strain distribution on the tissue layers causing injury \cite{nahum1994injury, ottenio2015strain}. Finally, if recoil occurs but not in the direction of the impact, again, shearing forces act on the skin, which are non-negligible\cite{zhang1994reaction}.

This is particularly relevant in unconstrained collision scenarios, where undesired contact is more likely than in constrained conditions, the direction of motion and recoil are rarely aligned unless the collision is expected and initiated by the human.

\section{Methodology} \label{sec:methodology}

In this section, we describe the experimental focus, conditions, setup, and procedure of the structured unconstrained impact experiments. 

\subsection{Research Objectives}

The questions we experimentally investigate with respect to safety in collisions with human hands are

\begin{itemize}
    \item[i)] Is it sufficient to consider perpendicular contact at collision for unconstrained collision in the low energy range for a conservative injury estimation based on the robot impact energy?
    \item[ii)] Is there a strong influence of other parameters besides the contact angle such as the human effective mass that needs to be considered?
\end{itemize}

As this is an exploratory experiment including a large number of different parameters, we focus on a descriptive analysis and do primarily not aim for statistical tests. We base our choice of human surrogates on previous work assessing injuries in collisions with the human hand \cite{Kirschner_2024_Towards, Kirschner_2024_unconst}. It is written that anatomical similarity between pig paws and human hands exists, but also the difference in skin thickness considering dermis and epidermis of pig skin (between 1 and 6 mm \cite{Shergold2006}, \cite{Summerfield2015}) and the human skin on the back of the hand ($\sim$ \SI{2.3}{mm} \cite{Oltulu2018}) is mentioned.

%Consequently, we use dissected pig dew claws and the back of pig feet as human finger and middle hand surrogates and investigate a distal and proximal collision location on each specimen as shown in Fig. \ref{fig:specimen}. 
%The observed injuries also correspond to the formerly presented injury schematics in \cite{Kirschner_ICRA24}, and are shown here again for the sake of completeness

\subsection{Impact Experiments}

In this study, we apply a replicate version of the pendulum-test stand mentioned in Sec. \ref{sec:State of the Art} and adapt it, such that we can assess shearing contact conditions. Otherwise, we aim to closely replicate the procedures to avoid variance in the results by design.

\subsubsection{Experimental Setup Design}

% \begin{table}[tpb]
%     \centering
%     \caption{Experimental settings}
%     \vspace{-0.1cm}
%     \begin{tabular}{p{11mm}p{13mm}p{27mm}p{18mm}}
%         \toprule
%         Parameter & & Value  & Based on   \\ 
%         \midrule
%         \multirow{2}{*}{\parbox{16mm}{$m$ [\SI{}{kg}]}} & impacted & $\sim$ \SI{2.6}{}, \SI{3.4}{} & hand/arm \cite{iso15066} \\
%          &impacting & $\sim$ \SI{3}{}, \SI{5}{}, \SI{7}{}, \SI{9}{} &  robot reflected inertia \cite{Kirschner_2021_Notion}\\
%          \parbox{16mm}{$v$ [\SI{}{m/s}]} &  & \SI{0.25}{}, \SI{0.5}{}, \SI{1.0}{}, \SI{1.5}{}, \SI{2.0}{} &  \cite{iso_10218-2}, \cite{franka_FR3_da}\\
%          \multirow{3}{*}{\parbox{16mm}{impactors}} & wedge (W) & prism \ang{90}, boned &  \cite{Haddadin.2012}\\
%           & edge (E)  & tetrahedron \ang{90}, boned &  \cite{Haddadin.2012}\\
%            & sheet (S) & width \SI{1.5}{mm}, boned &  \cite{casalino_2018}\\
%         \bottomrule
%     \end{tabular}
%     \label{tab:Experimental parameters}
%     \vspace{-0.3cm}
% \end{table} 

\begin{figure}[t]
\centering
\includegraphics[width=1\linewidth]{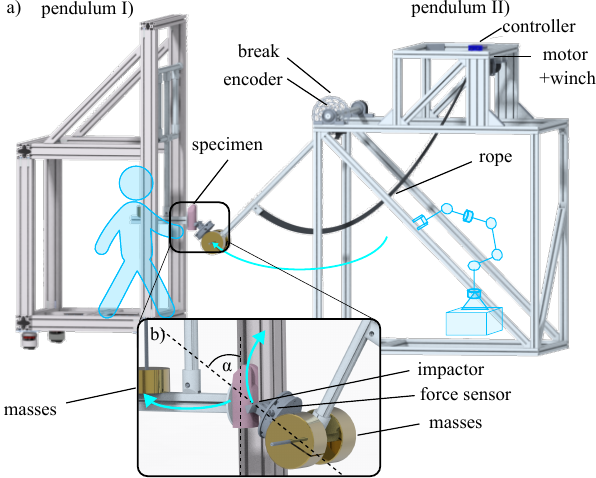}
\caption{Experimental setup for unconstrained shearing collision injury analysis.}
\vspace{-0.2cm}
\label{fig:setup}
\vspace{-2mm}
\end{figure}

The applied test stand is depicted in Fig. \ref{fig:setup}. The test rig models the features of the human body part (pendulum I) and the robot's effective mass and velocity (pendulum II). Both pendulums allow adapting their effective mass by adding weights.

For the first pendulum, we assume a) the human hand and lower arm are hit and the arm from distal to the elbow recoils freely, b) the human hand, lower arm, and upper arm are involved in the collision and the arm recoils with the rotation axis being inside the shoulder joint. We simulate these conditions based on the effective masses provided by ISO 10218-2  (hand: \SI{0.6}{kg}, lower arm: \SI{2}{kg}, upper arm: \SI{3}{kg}) \cite{iso_10218-2}. Consequently, we attach masses arranged in the line of motion. 

\begin{figure}[t]
\centering
\includegraphics[width=1\linewidth]{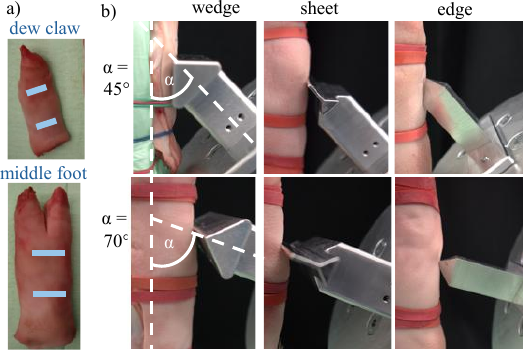}
\caption{Impact conditions: a) dissected specimen with collision locations and b) impact angle $\alpha$ for different impactors.}
\vspace{-0.3cm}
\label{fig:impact_angles}
\end{figure}

On top of this first, purely mechanical pendulum, the dew claws and middle foot of the pig front paws are tied to the impact location, adding $\sim$ \SI{0.3}{kg}. They are aligned, such that the chosen collision location from the ones is reached at the desired angle as depicted in Fig. \ref{fig:impact_angles}.

Pendulum-II, illustrated on the right side of Fig.~\ref{fig:setup}, represents the robot through its effective mass and supports the attachment of various contact geometries. The system is fully automated, using a compactRio platform from National Instruments for control and data acquisition at a sampling rate of \SI{2000}{Hz}. Its mechanical design includes a pendulum rod equipped with mounts for normed slit weights, allowing a total attached mass of up to \SI{8}{kg}. At the contact interface, a K3R110 force/torque sensor (ME Messsysteme) \cite{ME_datasheet_K3} is installed, which records force data at \SI{300}{Hz} via the GSV-8DS measurement amplifier \cite{gsv_datasheet}. Various impactors can be mounted on top of the force-torque (F/T) sensor. In this study, we apply the \SI{90}{^\circ} wedge, edge, and a \SI{1.6}{mm} thick sheet, fabricated from aluminum alloy EN AW-7075 with a hardness of \SI{150}{HB}. The pendulum’s angular velocity is measured using a TMCS-28-1k-KIT rotary encoder from ADI Trinamics \cite{encoder_datasheet}. A disc brake, actuated by a stepper motor, serves two purposes: (a) to position the pendulum at a desired height, and (b) to brake its motion post-impact, preventing secondary collisions due to backswing. The pendulum is lifted using a motorized belt and winch system to achieve a specific deflection angle $\alpha$, from which the desired impact velocity $v_{d}$ is determined using a calibration curve for the selected impact angle $\alpha$.

\begin{equation}
m_\mathrm{eff}=  \frac{J_{xx}^{\mathrm{(S)}}+m_\mathrm{p}l^2}{l_\mathrm{col}^2}  \, ,
\end{equation} 
is the pendulums effective mass where $J^{\mathrm{(S)}}_{\mathrm{xx}}$ is the inertia around the pendulum center of gravity, $m_\mathrm{p}$ the pendulum summed mass, $l $ the distance to the center of gravity, and $l_\mathrm{col}$ the distance to the point of collision \cite{Kirschner_2024_unconst}. 
 Table \ref{tab:meff} summarizes the effective masses applied in this study for pendulum-II based on the attached extra loads as depicted in Fig. \ref{fig:setup} with all impactors. For pendulum I, we apply a load such that we reach $\sim$\SI{2.6}{kg} (Ia) and $\sim$\SI{6.4}{kg} (Ib), which shall represent human hand with lower arm and the entire arms effective masses, in the range of the standard reported masses as listed above. % A few paragraphs earlier, it says: We simulate these conditions based on the effective masses provided by ISO/TS 15066:2016 \cite{iso15066} (\SI{0.6}{kg}, \SI{2}{kg}, \SI{3}{kg}). -> possibly redundant
 For both, pendulum Ia) and Ib), as well as all three impactors, we cause collisions with \SI{1.0}{m/s}, \SI{1.5}{m/s}, and \SI{2.0}{m/s} velocity at impact with all effective masses according to Table \ref{tab:meff}.

\begin{table}[t]
\centering
\caption{Effective mass ($m_{\text{eff}}$) values for different impactor types and added weights.}
\label{tab:meff}

\begin{tabular}{lccccc}
\toprule
\textbf{Impactor} & \textbf{0 kg} & \textbf{2×1 kg} & \textbf{2×2 kg} & \textbf{2×3 kg} & \textbf{2×4 kg} \\
\midrule
Wedge & 1.56 & 3.53 & 5.47 & 7.46 & 9.46 \\
Edge  & 1.43 & 3.40 & 5.64 & 7.33 & 9.33 \\
Sheet & 1.52 & 3.49 & 5.43 & 7.43 & 9.42 \\
\bottomrule
\end{tabular}
\vspace{-5mm}
\end{table}
% \begin{table}[]
%     \centering
%     \caption{Modelled pendulum effective masses}
%     \begin{tabular}{ccccccc}
    
%     \toprule
%         pendulum & load & $J_{xx}^{\mathrm{(S)}}$ & $l$ & $l_{col} $ & $m_\mathrm{p}$ & $m_\mathrm{p,eff} $  \\
%          & $[kg]$ & $[kgmm^2]$ & $[mm]$ & $[mm]$ & $[kg]$ & $[kg]$ \\
%          \midrule
%         I& 1 &  409,604.47 & 518 & 794 & 4.54 & 2.58 \\
%         I& 4 & 648,362.99 & 670 & 794 & 7.49 & 6.36 \\
%         %II & 0 & & & & & \\
%         II & 2 & 531,378.16 & 782 & 990 & 4.26 & 3.20 \\
%         II & 4 & 583,728.08 & 842 & 990 & 6.26 & 5.12 \\
%         II & 6 & 612,297.60 & 873 & 990 & 8.26 & 7.04 \\
%         II & 8 & 631,599.33 &  892 & 990 & 10.26 &  8.97\\
%          \bottomrule
%     \end{tabular}
    
%     \label{tab:eff_mass}
%     \vspace{0mm}
% \end{table}

\subsubsection{Experimental Protocol}

The experimental procedure reported in the main phases includes Preparation, Experimental Phase, Injury analysis, and Evaluation.

\subsection{Preparation Procedure}

Prior to testing, specimens underwent the following standardized preparation process. Specimens were thawed for approximately 12 hours at room temperature ($\sim$\SI{23}{^\circ C}). Then, dew claws were carefully dissected from the central region of the foot to isolate the target specimen, dew claws and back of the front paw representing human finger and back of the hand as shown in Fig. \ref{fig:impact_angles} a). To maintain tissue hydration, samples were stored under a moistened towel. Pendulums were assembled, and specimens were securely mounted to Pendulum I by strings in preparation for impact testing.

\subsection{Experimental Phase}

The impact testing protocol follows the sequence:

\begin{enumerate}
    \item \textbf{Pendulum Alignment:} Both pendulums are aligned to ensure accurate distance for the contact angle.
    \item \textbf{Elevation:} Pendulums are lifted by the motor and winch to the predetermined height, where a mechanical brake holds them in place.
    \item \textbf{Release Mechanism:} The motor unwinds the restraining rope, and the brake is disengaged to initiate motion.
    \item \textbf{Collision and Shearing Interaction:} Pendulum II collides at the specified angle of impact ($\alpha$) with the pig specimen and moves along the pig specimen surface as indicated in Fig. \ref{fig:setup} b), simulating a shearing force on top of the specimen. 
    \item \textbf{Recoil and Catch:} Following impact, both pendulums recoil and are subsequently caught by the test operator and brake system to prevent secondary collision.
\end{enumerate}

\subsection{Injury Analysis:}

The specimen is dismantled from the test  setup, carefully, ensuring no additional damage. To evaluate the extent of physical damage following the shearing contact, first, a macroscopic assessment of the skin for visible injuries is conducted. In cases where deeper damage was suspected, dissection was performed to examine potential lesions affecting tendons or bones. This is especially the case, if deeper skin cuts were present already. The observed injuries were categorized into four distinct classes: no visible damage, skin imprints (s-i), skin cuts (s-c), and tendon or bone (t/b) injuries.

\begin{figure}[t]
\centering
		\includegraphics[width=1\linewidth]{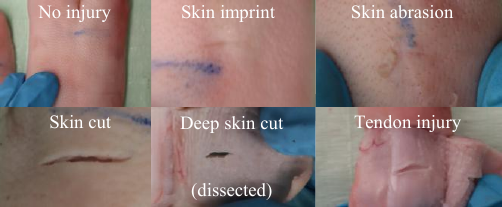}
		
	\caption{Observed injury types.}
	\label{fig:injuries}
	\vspace{-5mm}
\end{figure}

\subsection{Evaluation:}

To evaluate whether impact angle influences injury occurrence, we performed a Chi-Square test on a contingency table of injury status vs. collision angle. The hypotheses were defined as follows:
\begin{itemize} \item [H0:] Injury occurrence is independent of impact angle. \item[H1:] Injury occurrence depends on impact angle. \end{itemize}
The test statistic $\chi^2$ and $p$-value were computed in MATLAB 2022b using \textit{crosstab} with the degrees of freedom
\begin{equation} 
(r-1)(c-1) = 1 \, , 
\end{equation}
where $r$ and $c$ are the category counts of each variable.

To provide energy limits, we consider the entire moving mass of the pendulum representing the robot, as this is the maximum possible transferred energy and what we can efficiently limit by robot control. Thus, we provide the kinetic energy based on the measured impact velocity, $v_\mathrm{c}$, and the moving effective mass of the pendulum or drop test, $m_\mathrm{eff}$, as
\begin{equation}
K_\mathrm{max}=  \frac{m_\mathrm{eff}v_\mathrm{c}^2}{2}  \, ,
\end{equation} 
for all available studies. For each injury category, we generate box plots and use the lowest whisker (5th percentile) to describe the energy threshold that should not be crossed, such that the injury is prevented. Based on the final overview of energy thresholds, we decide whether the perpendicular or shearing contact in unconstrained collisions represents the more relevant threshold.

Due to the variance in force sensors applied among the studies, the comparison of forces is not ideal and, thus, neglected in this paper. %All studies apply different impactors to model the impact area of relevant tools those are also as in our study  a wedge \SI{90}{^\circ}, edge \SI{90}{^\circ} and the \SI{1.6}{mm} thick sheet 

\begin{table}[t]
\centering
\caption{Injury occurrences by collision angle across all experiments. The table highlights differences in injury types between 45° and 70° impact configurations.}
\renewcommand{\arraystretch}{1.2}
\setlength{\tabcolsep}{8pt}
\begin{tabular}{lccc}
\toprule
\textbf{Injury Classification} & \textbf{$\mathbf{\alpha}$ = \SI{45}{^\circ}} & \textbf{$\mathbf{\alpha}$ = \SI{70}{^\circ}} & \textbf{Total} \\
\midrule
Skin Imprint   & 154 & 255 & 409 \\
Skin Cut       & 27  & 80  & 107 \\
Skin Abrasion  & 31  & 19  & 50  \\
Tendon Injury  & 1   & 6   & 7   \\
\bottomrule
\end{tabular}

\label{tab:injury_comparison}
\vspace{-5mm}
\end{table}

\subsection{Assessing Injury Severity}

For a tractable risk assessment conforming to EN ISO 12100, the injury needs to be assigned into distinct severity classes. We perform the categorization into severity classes based on EN ISO 12100 \cite{iso12100}, and add the S0-category for no injury as suggested by \cite{iso_10218-2}.
    \begin{itemize}
     \item[S0]: no visible injury (e.g., subsiding pain, subsiding skin deformations)
     \item[S1]: minor reversible injury requiring first aid (e.g., epidermal and dermal skin lesion) 
     \item[S2]: reversible injury requiring professional medical attention (e.g., bone fractures, tendon, ligament, or muscle lesions)
     \item[S3]: irreversible injury (e.g., amputation of limbs, severe neurovascular injuries)
     \item[S4]: fatal injury (causing imminent death)
    \end{itemize}

This classification allows us to further propose thresholds for safe  interactions. In our experiments, S3- and S4-type injuries are not expected. For providing thresholds for safety, in this work, we argue, based on the state of the art (EN ISO 10218-2), that no injury is desired. %Please note that, based on a proper risk assessment, it may also be possible to set the thresholds differently if the frequency of occurrence of the probability of harm is very low.

%%%%%%%%%%%%%%%%%%%%%%%%%%%%%%%%%%%%%%%%%%%%%%%%%%%%%%%%%%%%%%%%%%%%%%%%%%%%%%%%
%%%%%%%%%%%%%%%%%%%%%%%%%%%%%%%%%%%%%%%%%%%%%%%%%%%%%%%%%%%%%%%%%%%%%%%%%%%%%%%%
\section{Results} \label{sec:results} 

Overall, in the shearing injury study for each combination of impactor, using pendulum Ia) or Ib) as human effective mass conditions (further referred to as Ia) and Ib)), collision angle $\alpha$, mass, and velocity, six experiments were carried out on pig dew claw and 3 on the pig front paw specimen. Providing a total of 1080 experiments. Here, we summarize the most important results.

\begin{figure}[t]
\centering
		\includegraphics[width=1\linewidth]{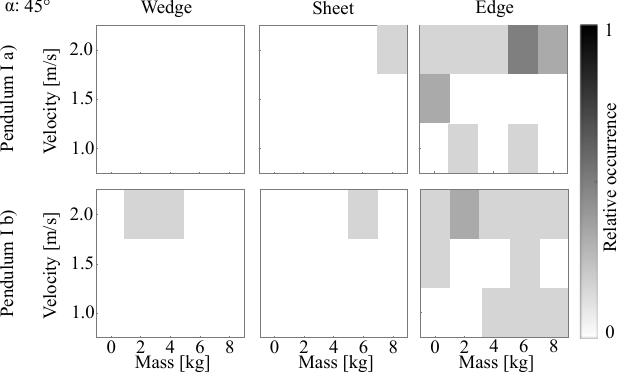}
		
	\caption{Observed percentage of injury for skin cuts occurring at \SI{45}{^\circ} impacts using pendulum Ia) and Ib) with \SI{2.6}{kg} and \SI{6.4}{kg} mass with all impactors.}
	\label{fig:45_probs}
	\vspace{-5mm}
\end{figure}

\subsection{Injury Type Occurrence}

We investigated the injury types, according to the description in Sec. \ref{sec:methodology} E. All injuries are exemplary shown in Fig. \ref{fig:injuries}. One additional injury type, in addition to the ones reported in the previous studies, was observed. As a result of the shearing nature of the experiment, skin abrasions, as depicted in Fig. \ref{fig:injuries} c), occurred. In the following, we report by injury type the percentage of the experiments in which they were observed. Table \ref{tab:injury_comparison} summarizes all observed injuries. %with  \textbf{$\mathbf{\alpha}$ = \SI{45}{^\circ}} 154 skin imprints, 31 skin abrasions, 27 skin cuts and one tendon injury were observed. For \textbf{$\mathbf{\alpha}$ = \SI{70}{^\circ}} it was 255 skin imprints, 19 skin abrasions, 80 skin cuts, and 6 tendon injuries.

We provide the most critical results graphically for the occurrence of skin cuts in Fig. \ref{fig:45_probs} for $\alpha =$ \SI{45}{^\circ} and Fig. \ref{fig:45_probs} for $\alpha =$ \SI{70}{^\circ}, where the term \textit{mass} refers to the added pendulum load. Each data point holds information from six experiments. A proportion of 1 (\SI{100}{\%}) indicates that the injury occurred in all six experiments, with varying shades of gray indicating different percentages.

\subsubsection{Skin Imprints (s-i)}

  \textbf{$\mathbf{\alpha}$ = \SI{45}{^\circ}:} For the wedge impactor, the percentage of s-i is very low in both Ia) and Ib), particularly at lower effective masses and velocities, where there is almost none. In contrast, with the sheet impactor, there is more s-i occurrence, with most conditions exceeding \SI{50}{\%} for Ia). For Ib), the collision condition with an \SI{8}{kg} extra weight and a velocity of \SI{2}{m/s} results in a s-i in \SI{100}{\%} of experiments. Also with the edge in Ia) and an \SI{8}{kg} extra load at a velocity of \SI{1.5}{m/s}, many s-i are reported.

 \textbf{$\mathbf{\alpha}$ = \SI{70}{^\circ}:}  For the wedge impactor with pendulum Ia), s-i occurrence is high in collisions with velocities above \SI{1}{m/s}, reaching even \SI{100}{\percent} in \SI{50}{\percent} of cases. At \SI{1}{m/s},  s-i occurrence is significantly lower, including one case without any s-i. Pendulum Ib) shows a similar distribution, with more observations at \SI{1}{m/s}. The sheet impactor generally exhibits lower  s-i occurrence across all conditions compared to the wedge. At \SI{1}{m/s}, only two cases showed any s-i. At higher velocities,  s-i occurrence increases but never reaches \SI{100}{\percent}. Pendulum Ib) with the same impactor shows a similar trend, with more s-i occurrence at lower velocities. For the edge impactor,  s-i occurrence is highest at \SI{1}{m/s}.

 \begin{figure}[t]
\centering
		\includegraphics[width=1\linewidth]{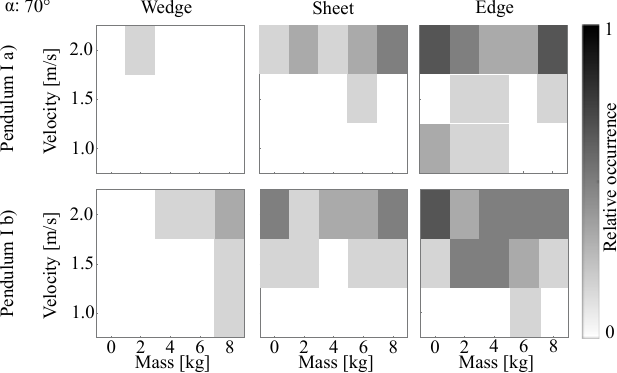}
		
	\caption{Observed percentage of skin cuts occurring at \SI{70}{^\circ} impacts using pendulum Ia) and Ib) with \SI{2.6}{kg} and \SI{6.4}{kg} mass with all impactors.}
	\label{fig:70_probs}
	\vspace{-5mm}
\end{figure}

\subsubsection{Skin Abrasion Injuries (s-ab)}

 \textbf{$\mathbf{\alpha}$ = \SI{45}{^\circ}:} For the wedge impactor with pendulum Ia), the s-ab percentage is high at velocities of $\geq$\SI{1.5}{m/s} independent of mass. At \SI{1.0}{m/s}, only one case of s-ab is reported. Using the pendulum, Ib) a similar pattern is observed, but without s-ab at \SI{1.0}{m/s}. The sheet impactor exhibits low numbers of s-ab in both scenarios, with more occurrences in Ia), generally concentrated at high velocities or low velocities and high mass. Uniquely, for the edge impactor, we observed that scenario Ib) shows more s-ab than Ia), where in Ia) s-ab occurred with the edge impactor.
 
 \textbf{$\mathbf{\alpha}$ = \SI{70}{^\circ}:}  S-ab is reported once each with the wedge and sheet impactor. In contrast, s-ab incidence with the edge is higher across all test conditions, with probabilities typically ranging from \SI{30}{\percent} to \SI{40}{\percent}.
 
 \subsubsection{Skin Cut Injuries (s-c)}
 
 \textbf{$\mathbf{\alpha}$ = \SI{45}{^\circ}:} As depicted in Fig. \ref{fig:45_probs}, for the wedge impactor, the no s-c occurred for case Ia), while a few incidents occurred at \SI{2}{m/s} using setting Ib). The sheet impactor shows a similar trend, with two s-c observed at \SI{2}{m/s} for Ia) and Ib), respectively. In contrast, with the edge impactor, a more distributed s-c occurrence is reported. Specifically, s-c incidents are observed across all conditions at \SI{2}{m/s}, and also at lower velocities, though without a notable consistent pattern.

\begin{figure}[t]
\centering
		\includegraphics[width=1\linewidth]{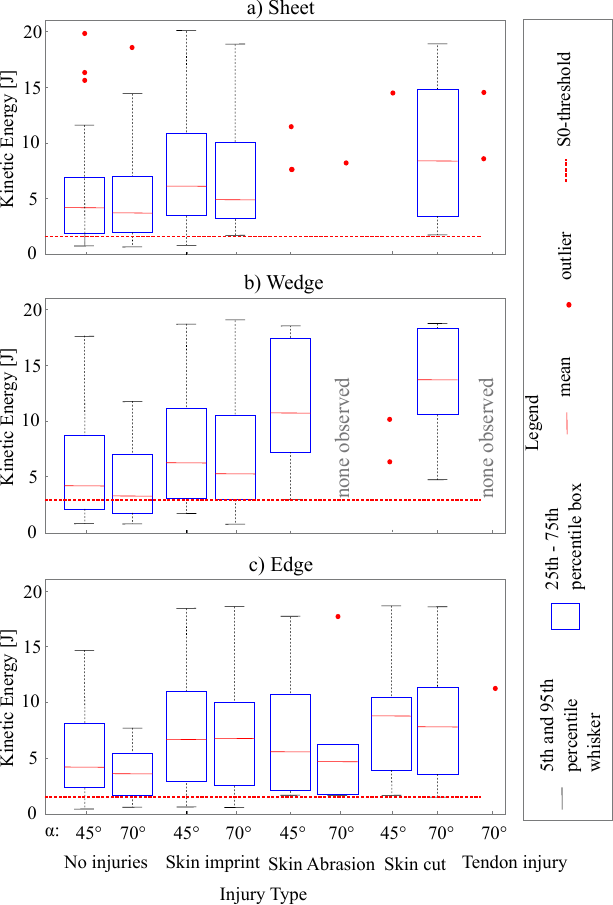}
		
	\caption{Energy distribution of observed injury types using the respective impactors for Ib).}
	\label{fig:energy_results_Ib}
	\vspace{-5mm}
\end{figure}

 \textbf{$\mathbf{\alpha}$ = \SI{70}{^\circ}:}  As depicted in Fig. \ref{fig:70_probs}, using the wedge in Ia) only one condition with s-c occurred, at \SI{2}{m/s} with a low mass. For Ib) in all conditions of the highest mass s-c incidents are reported, and in addition to that, two other cases at \SI{2}{m/s} and medium to high mass. The sheet impactor for Ia) causes s-c in all cases with \SI{2}{m/s} speed, with varying occurrence percentage, and one case is reported at \SI{1.5}{m/s} with medium mass. Using the pendulum Ib) for the sheet impactor shows an even higher probability for s-c occurrence. At \SI{2}{m/s}, all different weight configurations result in s-c injuries. With \SI{1.5}{m/s} collision speed only for one, medium-weight, there are s-c injuries present. For the edge impactor s-c occurred over all conditions frequently, especially for velocities above \SI{1.5}{m/s}.
 With every velocity, s-c was observed. In Ib), s-c occurrence with \SI{1.5}
{m/s} and \SI{2}{m/s} is comparable. At \SI{1.0}{m/s} only one s-c occurred with \SI{8}{kg} mass.

  \subsubsection{Tendon Injuries (t-i)}
  \textbf{$\mathbf{\alpha}$ = \SI{45}{^\circ}:} There one t-i reported using the edge impactor and pendulum Ib) at \SI{2}{m/s}.
  
\textbf{$\mathbf{\alpha}$ = \SI{70}{^\circ}:} No bone or tendon injuries were observed with the wedge impactor. One case of tendon injury occurred with the sheet impactor on the hand at \SI{2}{m/s} and relatively low weight. Collisions with the pendulum Ib) with the sheet impactor resulted in two tendon injuries, both at the same weight configuration (\SI{7.4}{\kilogram}) and velocities of \SI{1.5}{m/s} and \SI{2}{m/s}. For the edge impactor, two cases were observed in the first scenario at medium and high velocities with medium weight, and one case in the second scenario at \SI{2}{m/s}.

\subsection{Influence of the impact angle}
The calculated injury probabilities for each impact angle were:
\[
p_{\text{injury},\,45^\circ} = 0.11, \quad p_{\text{injury},\,70^\circ} = 0.19
\]
The Chi-Square test showed the following result:
\[
\chi^2 = 16.73, \quad p < 0.0001
\]

Since the p-value is below the significance level $\alpha = 0.05$, the null hypothesis of independence was rejected. This indicates a statistically significant association between impact angle and injury probability.

 \begin{figure}[t]
\centering
		\includegraphics[width=1\linewidth]{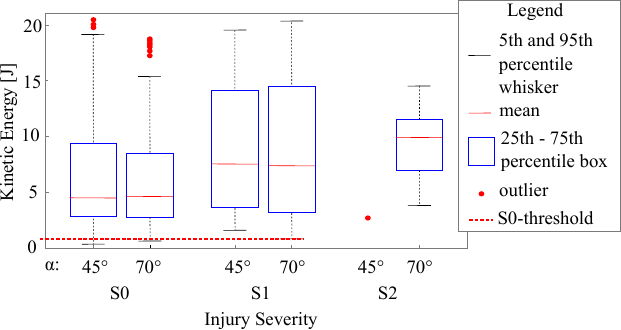}
		
	\caption{Observed Injury severity in unconstrained, shearing contacts over all impactors distributed over the kinetic energy of the pendulum in the contact.}
	\label{fig:energy_results}
	\vspace{-5mm}
\end{figure}

\subsection{Energy Causing Injuries for Unconstrained Shearing Collisions}

To assess the relevant energy bounds for collision scenarios, we investigate the energy distribution under which every injury type was observed, as depicted exemplarily in Fig. \ref{fig:energy_results_Ib} for the  experiment condition Ib. The impact condition represents the most severe injury cases reported in this study. 

No skin cuts or abrasions indicate the thresholds for the injury severity level S0. With the sheet, we observe no skin injuries up to \SI{1.10}{J}, depicted in Fig. \ref{fig:energy_results_Ib} a). Fig. \ref{fig:energy_results_Ib} b) shows that with the wedge impactor, we observe no skin cuts or abrasions until \SI{2.68}{J}. With the unprocessed edge, we observe for both angles no skin cuts or abrasions until a value of \SI{0.78}{J}, as depicted in Fig. \ref{fig:energy_results_Ib} c).

 Combining all the experimental data for the different impactors and hand or arm scenario differentiated by the impact angles creates the plot in Fig. \ref{fig:energy_results}. In this plot, the injuries are categorized based on the required medical treatment: S0 for no medical treatment, S1 for first aid, and S2 for medical treatment. The trend of more severe injuries occurring at higher kinetic energies is evident.
 
 \textbf{$\mathbf{\alpha}$ = \SI{45}{^\circ}:}
 For no injury the median energy is \SI{4.54}{J} with the 25th percentile being at \SI{2.82}{J} and the 75th
 percentile at \SI{9.40}{J}. S1 injuries have the median at \SI{7.57}{J} with the 35th percentile at \SI{3.67}{J} and
 the 75th percentile at \SI{14.23}{J}. The injury S2 is not presented in the Fig. \ref{fig:energy_results} as it only contains one data point at \SI{2.87}{J}, which cannot be represented as a distribution. The safety threshold here is defined as the minimal value of the S1 injuries. The value here is at \SI{1.61}{J}. No S2 injuries were observed, such that also no threshold to distinguish between the S1 and S2 injury can be provided.
  
 \textbf{$\mathbf{\alpha}$ = \SI{70}{^\circ}:} For S0 injuries, the median kinetic energy is \SI{4.71}{J}, with the 25th percentile at \SI{2.97}{J} and the 75th percentile at \SI{8.63}{J}. The median for the S1 injury is at \SI{7.81}{J}, with the 25th percentile at
 \SI{3.89}{J} and the 75th percentile at \SI{14.66}{J}. The most severe injuries, S2, show a median of \SI{9.93}{J} with a 25th percentile of \SI{6.99}{J} and 75th percentile of \SI{11.51}{J}. The safety threshold here is
 at \SI{0.78}{J} for the S1 injury when calculated using the minimal value for this injury and at \SI{3.78}{J} for the S2 injury.

\begin{figure}[t]
\centering
		\includegraphics[width=1\linewidth]{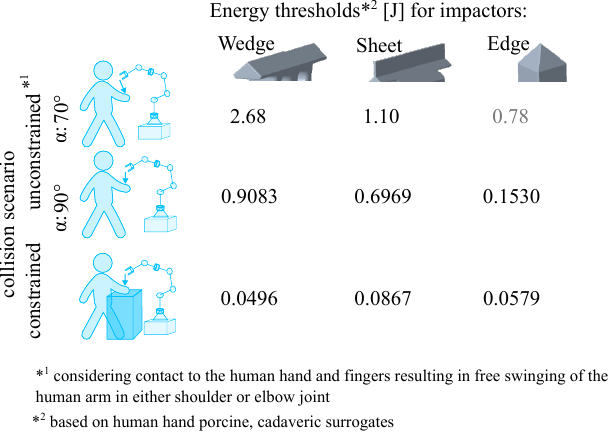}
		
	\caption{Injury Protection Database (surrogate studies): Most conservative thresholds for avoiding occurrence of any open skin injury from surrogate studies on pig front paws.} % why 0.78 in grey an why without alpha = 45°
	\label{fig:energy_thresholds}
	\vspace{-5mm}
\end{figure}

 \section{Overview of Energy-density limits for safe collisions}
To apply safety-relevant limits for safe energy-based control, we summarize the energy thresholds resulting from three studies for sharp, unprocessed impactors on the human hand and fingers. Fig. \ref{fig:energy_thresholds} summarizes all resulting energy thresholds under the considered conditions. As indicated in the previous subsection and visible from Fig. \ref{fig:45_probs} and \ref{fig:70_probs}, there was a low number of injury events (skin cuts) still occurring even at the lowest tested mass and velocity settings. This indicates that the threshold for injury for edges in unconstrained shearing contacts requires verification by more experiments with less energetic collisions.

Please note that, as injury occurs based on energy density, not energy alone, these thresholds only apply if the contact surface is comparable. Additionally, these values are the 5th percentile. Injury is not fully excluded. Generally, EN ISO 12100 advises being moderately conservative and rather assess injury risk as higher than too low \cite{iso12100}.

\section{Discussion} \label{sec:Discussion}
The shearing experiments reveal a clear trend: kinetic energy correlates with injury severity, as expected, and collisions at larger angle and with higher effective mass on the human side result in more injury cases. The reporting of forces was omitted in this paper, as yet no conclusive results could be taken from these. Force distribution in the contact direction among all contact types, and also plotted over occurring injuries, did not show any interpretable patterns. This may be a result of the short contact duration.  Notably, skin abrasions and skin cuts exhibit closely aligned energy values, making it difficult to define a distinct boundary between these injury types. Among the tested angles, the \SI{90}{^\circ} (perpendicular) collisions appear to be the most severe. When comparing the results from \cite{Kirschner_2024_unconst} with the new study on unconstrained shearing contacts, we observe that shearing scenarios allow for more than twice the energy threshold compared to perpendicular impacts. This suggests that for the conservative injury estimation in unconstrained human hand and finger collision scenarios, assessing perpendicular contact is required \footnote{This may not apply to heavier body parts with larger effective mass}.

\begin{figure}[t]
\centering
		\includegraphics[width=1\linewidth]{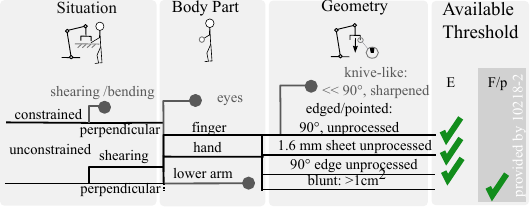}
	\caption{Scenario-based threshold selection for choosing safety threshold focused on human hand and fingers.}
	\label{fig:dec_matrix}
	\vspace{-5mm}
\end{figure}

To ensure safe controller deployment across real-world scenarios, decision matrices such as the one shown in Fig.~\ref{fig:dec_matrix} must be considered. The entire control pipeline must transparently enforce low energy thresholds, as indicated in Fig.~\ref{fig:energy_thresholds}, particularly for short-duration initial impacts. This assumes that no injury, even a minor one, is acceptable. However, if the robotics community begins to tolerate reversible injuries like superficial skin cuts under low-likelihood conditions, this perspective may evolve.

In line with current safety standards, this and related studies do not differentiate skin cuts by depth. Even superficial epidermal injuries, such as paper cuts, are considered unacceptable. While human tissue has not yet been tested under these scenarios, future studies should reassess injury thresholds for human hands and fingers based on the presented values. Given the robustness of porcine tissue, it is reasonable to assume that human thresholds will not be higher. Due to the chemical and structural similarity of porcine and human epidermis, we expect comparable injury thresholds, particularly for S0-level injuries.

\section{Conclusion} \label{sec:Conclusion}
%%%%%%%%%%%%%%%%%%%%%%%%%%%%%%%%%%%%%%%%%%%%%%%%%%%%%%%%%%%%%%%%%%%%%%%%%%%%%%%%
This study expands existing injury protection databases by investigating unconstrained shearing contact scenarios using porcine surrogates as a human hand and finger substitute. Across 1080 experiments, we observed that the collision angle significantly affects injury probability, with a Chi-Square test confirming a statistically significant association between impact angle and injury occurrence. Shearing contact consistently allowed for higher energy thresholds before injury onset compared to perpendicular impacts. By establishing an energy threshold across impactor geometries and contact scenarios, we provide the first comprehensive energy-based injury protection dataset relevant to human-robot interaction. These thresholds can be directly applied to energy-limiting controllers, enabling safe interaction across a wider range of realistic collision scenarios, including those involving sharp or pointed objects. By future validation with human experiments, additional experiments considering more angles for shearing contacts, and more detailed statistical evaluations to predict injury thresholds well justified safety standards can be achieved and finally robotic systems may become safe to deploy in the real world among humans. 

%Future work should validate these results with cadaveric, and we are able to refine energy-density limits for more complex conditions that they can integrate into standardized risk assessment frameworks. Eventually, these efforts will allow for the development of robust systems that support the safe development of collabortaive robots in real-world environments.    

 \section*{ACKNOWLEDGMENT}
% Omitted for anonymous review
 
% \vspace{-0mm}
 We gratefully acknowledge the funding of the Lighthouse Initiative KI.FABRIK Bayern Phase 1: Aufbau Infrastruktur by the Bavarian State Ministry for Economic Affairs, Regional Development and Energy (StMWi), the Lighthouse Initiative Geriatronics by LongLeif GaPa gGmbH (Project Y) and of the Bavarian State Ministry for Economic Affairs, Regional Development and Energy (StMWi) as part of the project SafeRoBAY (grant number: DIK0203/01).
%  \vspace{-1mm}

%%%%%%%%%%%%%%%%%%%%%%%%%%%%%%%%%%%%%%%%%%%%%%%%%%%%%%%%%%%%%%%%%%%%%%%%%%%%%%%%

\bibliographystyle{IEEEtran} %unsrt}
\bibliography{literature.bib}

\end{document}